\documentclass[letterpaper, 10 pt, conference]{ieeeconf}  

\IEEEoverridecommandlockouts                              

\usepackage{times}

\usepackage{graphicx}
\usepackage{amsmath}
\usepackage{amssymb}
\usepackage{booktabs}
\usepackage{graphics}
\usepackage{epsfig}
\usepackage{float}
\usepackage{cite}
\usepackage{multirow}
\usepackage{svg}
\usepackage{algorithm}
\usepackage{algpseudocode}
\usepackage{import}
\usepackage{subcaption}

\usepackage[pagebackref=true,breaklinks=true,letterpaper=true,colorlinks,bookmarks=false]{hyperref}

\usepackage[capitalize]{cleveref}
\crefname{section}{Sec.}{Secs.}
\Crefname{section}{Section}{Sections}
\Crefname{table}{Table}{Tables}
\crefname{table}{Tab.}{Tabs.}

\newcommand{\etal}{\textit{et al.}}

\title{\LARGE \bf
Multi-Task Consistency for Active Learning
}

\author{Aral Hekimoglu\\
Technical University of Munich\\
{\tt\small aral.hekimoglu@tum.de}
\and
Philipp Friedrich\\
BMW Group\\
{\tt\small philipp.pf.friedrich@bmw.de}
\and
Walter Zimmer\\
Technical University of Munich\\
{\tt\small zimmer@cit.tum.de}
\vspace{0.5cm}
\and
~~~~~Michael Schmidt\\
~~~~~BMW Group\\
{~~~~~~~\tt\small michael.se.schmidt@bmw.de}
\vspace{0.5cm}
\and
~~Alvaro Marcos-Ramiro\\
~~~~BMW Group\\
{~~~~~~\tt\small alvaro.marcos-ramiro@bmw.de}
\vspace{0.5cm}
\and
~~~Alois Knoll\\
~~~~~Technical University of Munich\\
{~~~~\tt\small knoll@cit.tum.de}
}

\begin{document}

\maketitle
\thispagestyle{empty}
\pagestyle{empty}

\begin{abstract}

Learning-based solutions for vision tasks require a large amount of labeled training data to ensure their performance and reliability. In single-task vision-based settings, inconsistency-based active learning has proven to be effective in selecting informative samples for annotation. However, there is a lack of research exploiting the inconsistency between multiple tasks in multi-task networks. To address this gap, we propose a novel multi-task active learning strategy for two coupled vision tasks: object detection and semantic segmentation. Our approach leverages the inconsistency between them to identify informative samples across both tasks. We propose three constraints that specify how the tasks are coupled and introduce a method for determining the pixels belonging to the object detected by a bounding box, to later quantify the constraints as inconsistency scores. To evaluate the effectiveness of our approach, we establish multiple baselines for multi-task active learning and introduce a new metric, mean Detection Segmentation Quality (mDSQ), tailored for the multi-task active learning comparison that addresses the performance of both tasks. We conduct extensive experiments on the nuImages and A9 datasets, demonstrating that our approach outperforms existing state-of-the-art methods by up to 3.4\% mDSQ on nuImages. Our approach achieves 95\% of the fully-trained performance using only 67\% of the available data, corresponding to 20\% fewer labels compared to random selection and 5\% fewer labels compared to state-of-the-art selection strategy. Our code will be made publicly available after the review process.

\end{abstract}

\section{Introduction} \label{sec:intro}
Object localization and classification are critical for planning and executing safe and comfortable autonomous driving. Recent deep learning methods have demonstrated state-of-the-art (SOTA) performance on 2D object detection \cite{liu2022swin, xu2021end, duan2019centernet} and semantic segmentation \cite{yuan2020object, cheng2020panoptic} tasks. However, achieving high accuracy in these tasks comes at a high computational cost when handled separately, making them unsuitable to be used together for real-time autonomous driving. To address this challenge, multi-task learning has emerged as a promising solution. By sharing computations between related tasks, multi-task learning can achieve high accuracy while meeting real-time requirements. Recent publications showed that networks that predict both 2D object detection, and pixel-wise semantic segmentation perform better on both tasks compared to the single-task trained networks \cite{dvornik2017blitznet, ebert2022multitask}. In this paper, we focus on the problem of multi-task active learning in autonomous driving, aiming to maximize performance across multiple tasks while minimizing the need for large amounts of labeled training data.

\begin{figure}[t]
    \begin{center}
    \includegraphics[width=0.99\linewidth]{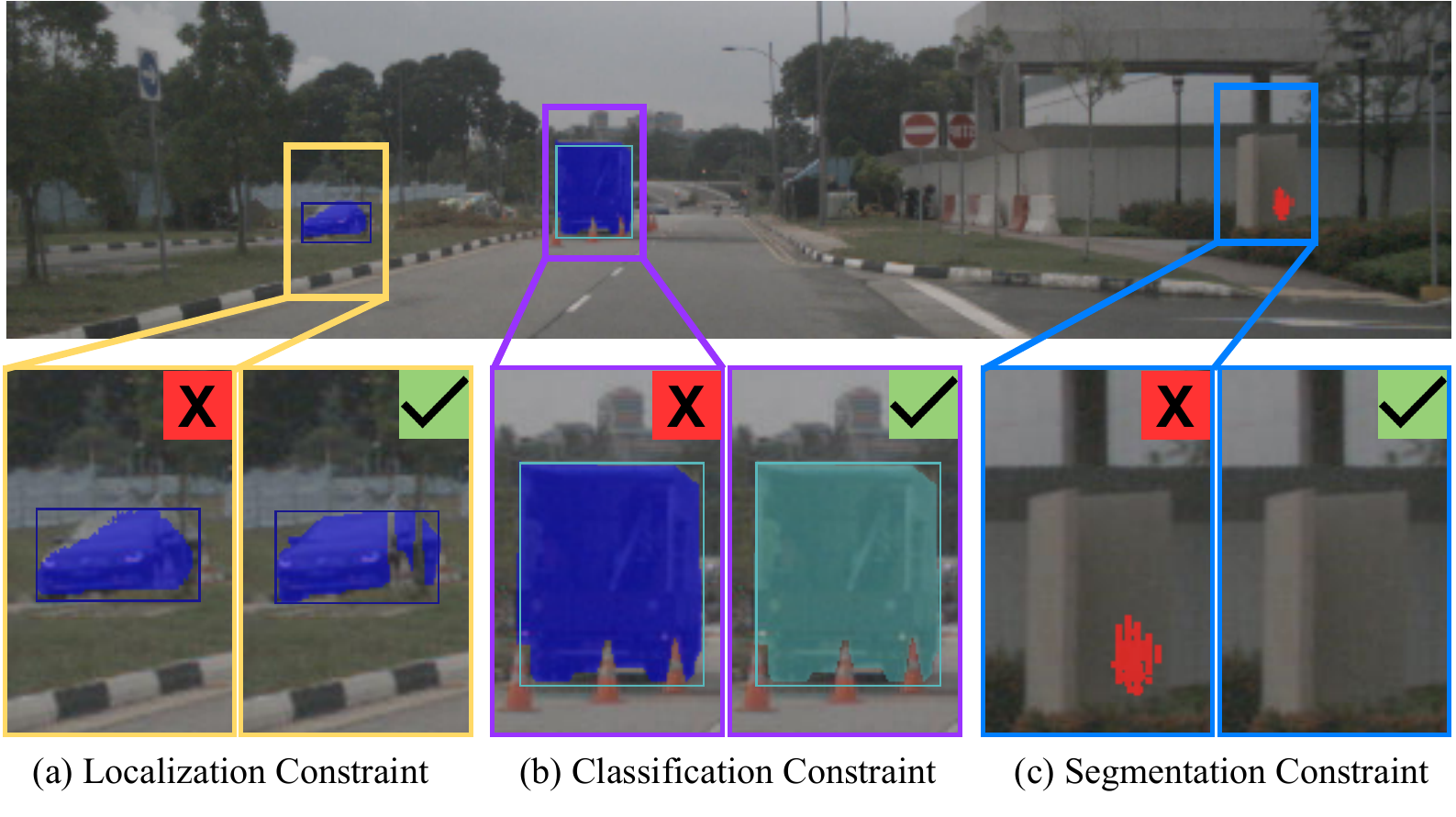}
    \end{center}
    \caption{Consistency constraints between object detection and semantic segmentation. (a) Mask covers all pixels of a detected object. (b) A detected object and the segmentation mask that covers it share the same predicted class distribution. (c) No pixels outside of the detected boxes are segmented with an object class.}
    \label{fig:intro_overview}
\end{figure}

\begin{figure*}[t]
    \begin{center}
    \includegraphics[width=1.0\linewidth]{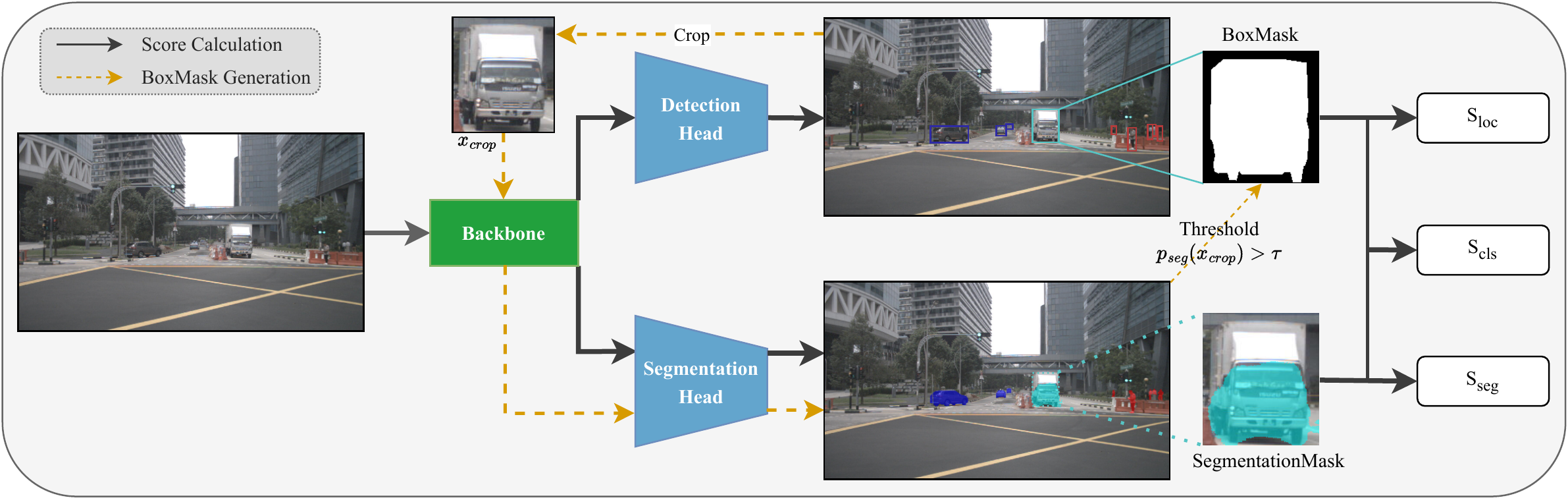}
    \end{center}
    \caption{The proposed inconsistency-based selection strategy. Detection boxes and the segmentation mask are obtained from a multi-task network architecture consisting of a backbone and two task heads. The BoxMask is generated by cropping the region around the detected box, passing it through the backbone and segmentation head, and applying a threshold $\tau$ to the class probability corresponding to the class of the detected box. Our proposed strategy focuses on three inconsistency scores by using the BoxMask and segmentation mask.}
    \label{fig:method_overview}
\end{figure*}

Active learning (AL) \cite{elezi2022not, colling2020metabox} is a technique for selecting the most informative samples for training a machine learning model under labeling budget constraints. In a typical single-task active learning loop, the model's predictions on the remaining unlabeled data are used to identify the samples that would be most beneficial for further training. In vision-based single-task settings, AL has been used to improve performance in tasks such as image classification \cite{beluch2018power, yoo2019learning}, object detection \cite{roy2018deep, desai2019adaptive, tang2021towards, li2021deep, choi2021active, hekimoglu2022efficient}, and semantic segmentation \cite{huang2021semi, golestaneh2020importance}. One effective approach is inconsistency-based selection, which identifies samples for annotation by measuring the inconsistency of the model's predictions across different augmentations of the input data. For instance, CALD \cite{yu2022consistency} explores various augmentations for object detection, and EquAL \cite{golestaneh2020importance} measures inconsistency between an image and its flipped version for semantic segmentation. However, to our knowledge, no existing work has explored the use of inconsistency between multiple tasks in the context of multi-task active learning.

Our novel strategy incorporates the concept of inconsistency-based selection from active learning and applies it to multi-task learning by leveraging the inconsistency between two coupled vision tasks, namely 2D object detection and semantic segmentation. Our approach quantifies the inconsistency between these two tasks to identify informative samples across both of them to maximize the performance while minimizing the amount of labeled data needed for training.

To measure the inconsistency between 2D object detection and semantic segmentation, we define three constraints that specify how the two tasks are coupled together. The first constraint requires that the segmentation mask covers all pixels of the detected objects (\cref{fig:intro_overview}a). The second constraint requires that a detected object and the segmentation mask that covers it share the same predicted class distribution (\cref{fig:intro_overview}b). The third constraint requires that no pixels outside the detected boxes are segmented with an object-class (\cref{fig:intro_overview}c). To map the object detection predictions to a similar pixel-wise output as the semantic segmentation predictions, we define BoxMask to identify the pixels belonging to the object detected by a bounding box. BoxMask enables us to quantify the three constraints by measuring the overlap between the object detection and semantic segmentation predictions. Based on these constraints, we propose three scores utilizing the BoxMask that quantify the inconsistency between the two coupled tasks.

Our main contributions are the following:
\begin{itemize}
    \item A novel multi-task active learning strategy that effectively leverages the inconsistency between 2D object detection and semantic segmentation to improve performance on both tasks and reduce the amount of labeled data needed for training.
    \item A novel method for identifying the pixels belonging to a detected object (\textit{BoxMask}) and using it to quantify the constraints between two tasks into selection scores.
    \item A comprehensive qualitative and quantitative comparison of the proposed approach and multi-task active learning baselines against SOTA baselines that are outperformed by up to 3.4\% mDSQ and
    5\% in data savings rate.
\end{itemize}

\section{Related work}
\subsection{Active learning}

AL methods for object detection measure uncertainty of a detected box through either classification or localization uncertainty \cite{brust2018active, aghdam2019active, elezi2022not, yuan2021multiple, kao2018localization, choi2021active}. Recent methods \cite{elezi2022not, yu2022consistency} leverage inconsistency between the predictions of the network when given different augmented versions of the sample to define the robustness of a sample and select samples that are less robust. For example, Elezi et al. \cite{elezi2022not} use horizontal flipping, while Yu et al. \cite{yu2022consistency} explore various augmentations to obtain multiple outputs and measure the inconsistency to define a selection score. Once the uncertainty of an object is estimated, the scores of all detections are aggregated using either the sum, the average, or the maximum of the scores, and the resulting image score is used to rank the images for annotation.

AL methods for semantic segmentation also utilize inconsistency methods to define uncertainty \cite{golestaneh2020importance, siddiqui2019viewal, li2020uncertainty, huang2021semi}. For instance, Golestaneh et al. \cite{golestaneh2020importance} apply horizontal flipping to the image and measure the inconsistency through the KL-divergence of the predictions from the original and the flipped image. The unit of data queried in AL methods for semantic segmentation varies, with methods querying either whole images \cite{golestaneh2020importance, xie2020deal} or regions \cite{siddiqui2019viewal, casanova2020reinforced, li2020uncertainty, kasarla2019region, xie2022towards}. In our work, we choose to query whole images for labeling since we are also interested in obtaining object detection labels. Notably, our approach is the first to utilize inconsistency between predictions of different tasks to define multi-task uncertainty and use it for AL selection.

Learning Loss is a task-agnostic strategy proposed by Yoo \etal \cite{yoo2019learning}, using a loss prediction module. The network learns to predict the target loss for unlabeled inputs, and samples with the highest predicted loss are selected for labeling. As this approach is task-agnostic, it can be adapted for multi-task networks and serves as a relevant comparison in our work.

Diversity-based methods aim to ensure a diverse training set that covers the input space. One such method is the use of a core-set, as proposed by Sener et al. \cite{sener2018active}, where diversity is defined as the Euclidean distance between intermediate network features for each image. Another method, CDAL, proposed by Agarwal et al. \cite{agarwal2020contextual}, exploits contextual diversity with respect to the predicted classes, and has been applied to object detection and semantic segmentation. These methods are applicable to the multi-task active learning scenario and provide additional baselines for comparison.

\subsection{Multi-task learning}

Multi-task learning has been studied extensively, and readers are referred to a survey by Crawshaw \etal \cite{crawshaw2020multitask}. In multi-task architectures, the hidden layers of a backbone model are shared among different tasks, and have separate heads that predict each task \cite{zhao2018modulation, liu2019end, ghiasi2021multitask}. Multi-task optimization deals with the joint objective function in multi-task settings, such as how to weigh losses of individual loss functions for different tasks \cite{gong2019comparison, kendall2017multitask, liu2019end, liu2019multi}. Kendall \etal \cite{kendall2017multitask} explored weighting each loss by its corresponding single-task uncertainty, using homoscedastic uncertainty for weighing the multi-task loss.

The idea of joint semantic segmentation and object detection was first investigated for shallow networks in \cite{hariharan2014simultaneous, papadopoulos2017extreme, ren2015faster}. These studies demonstrated that learning both tasks simultaneously can be better than learning them independently. Salscheider \etal \cite{salscheider2020simultaneous} employed a shared backbone and heads in their work, and we adopted this approach as our multi-task network.

\subsection{Multi-task active learning}

To our knowledge, no prior research has investigated multi-task active learning (MTAL) for object detection and semantic segmentation. However, active learning has been successfully applied to multi-task settings in other domains, such as Natural Language Processing (NLP). Reichart and Rappoport \cite{reichart2008multi} proposed alternating two single-task focused data selection strategies in each cycle, while Ikhwantri et al. \cite{ikhwantri2018multi} randomly selected a task for each cycle. Instead of alternating between two single-task scores, we propose a novel approach that generates a single score for selecting interesting samples relevant to both tasks.

\section{Methodology} \label{sec:methodology}
\subsection{MTAL problem overview}\label{sec:problem_overview}

The goal of this work is to tackle the AL problem of iteratively selecting samples from a large pool of unlabeled data $X^U$ to be labeled by an oracle, to improve the performance of a multi-task object detection and semantic segmentation network. Specifically, we consider each sample $(x,y_{det}, y_{seg})$ as a triplet, where $x$ is an image, $y_{det}$ is the set of objects, and $y_{seg}$ is the pixel-wise segmentation label belonging to one of the semantic classes $C_{seg}$. Detection labels $y_{det}$ consists of bounding box coordinates $(y_{box})$ and corresponding categories $(y_{cls})$ belonging to one of the object classes $C_{det}$, where $C_{det} \subseteq C_{seg}$.

Our multi-task network consists of a shared backbone and two single-task heads, as shown in \cref{fig:method_overview}. The network predicts object boxes $p_{det}$ consisting of $(p_{box}, p_{cls})$ and a segmentation mask $p_{seg}$ for each input image $x$. In each AL cycle, the network is trained on the labeled data $(X^L,Y^L)$, and a subset $S$ of unlabeled samples is selected for labeling. In the next cycle, the selected samples are added to the labeled dataset, and the network is trained again on the updated labeled dataset.

\subsection{Method overview}\label{sec:method_overview}

Object detection and semantic segmentation are two interconnected tasks shown to benefit from each other when combined \cite{dvornik2017blitznet}. The predictions from both tasks are inherently coupled, as objects detected in the former should align with the labeled regions in the latter. We propose an AL selection strategy that identifies samples where either task fails. To achieve this, we measure the inconsistency between the predictions of object detection and semantic segmentation. These inconsistent areas indicate potential points of failure for both tasks and, as such, are deemed interesting for further labeling.

To this end, we define three constraints between the tasks to formulate a selection score as illustrated in \cref{fig:method_overview}.
\begin{enumerate}
  \item The segmentation mask should cover all pixels belonging to the detected object, ensuring that the entire object is accurately segmented for the given class. (\cref{sec:boxmask_localization})
  \item The segmentation mask and the detected object should have consistent class distributions. (\cref{sec:boxmask_classification})
  \item There should be no segmented pixel belonging to a class from the object detection outside the predicted bounding boxes. (\cref{sec:inverse_boxmask})
\end{enumerate}

\subsection{BoxMask generation strategy}\label{sec:boxmask_generation}

To quantify the constraints, we define a binary segmentation mask, BoxMask, that covers all pixels within a detected box belonging to the class of the detected object. A perfect BoxMask covers each pixel of the entire object of interest. \cref{fig:method_overview} illustrates our BoxMask generation strategy. We begin by cropping a region of the image around the detected box, and then pass it through the network. Using the segmentation head of our multi-task network, we generate a segmentation mask for the cropped region $p_{seg}(x_{crop})$. BoxMask is then defined as a binary mask where the predicted class probability in the new segmentation label is above a threshold $\tau$ for the class of the detected object.

\subsection{Localization consistency}\label{sec:boxmask_localization}

Our localization-focused inconsistency score measures the alignment between a detected object and its corresponding segmentation mask. To ensure consistent detection and segmentation, the segmentation mask within the detected box should cover the entire detected object without any missing regions.

We define the localization consistency score as the number of pixels in the predicted segmentation mask that match the corresponding pixels in the BoxMask for that detection. To account for varying object sizes and scales, we normalize the score by $|BM|$, the number of pixels in the BoxMask, resulting in a scale-invariant score. Mathematically, the localization consistency score is given by the following equation:

\begin{equation}\label{eq:boxmask_localization}
S_{loc} = \frac{1}{|BM|}\sum_{i, j \in BM}{\mathbb{I}{(p_{seg}(i,j)==c)}}
\end{equation}

where, $\mathbb{I}$ represents the indicator function, $c$ denotes the predicted class of the detected object, and $i,j$ represent the pixel coordinates in the BoxMask ($BM$).

\subsection{Classification consistency}\label{sec:boxmask_classification}

In \cref{sec:boxmask_localization}, we addressed the localization inconsistency between the BoxMask and the segmentation mask. However, this approach treats classes that are very different, such as \textit{Truck} and \textit{Pedestrian}, the same as classes that are more similar, such as \textit{Truck} and \textit{Bus}. To account for this, we propose a classification inconsistency score that considers the predicted class probabilities during the sample score calculation.

To achieve this, we transform the predicted object class distribution $p_{cls}$ into the same probability domain as the class probabilities predicted by semantic segmentation. Specifically, we set the probability of the classes not trained in object detection, i.e., $C_{seg} - C_{det}$, such as \textit{Road} and \textit{Sky}, to a negligible value to ensure that both tasks have the same number of classes, without impacting the score calculation. Since pixels within BoxMask should not contain any classes from $C_{seg} - C_{det}$, we consider this a valid assumption. The transformed probability vector, denoted as $\tilde{p}_{cls}$, contains the same classes as the segmentation task ($C_{seg}$).

The classification inconsistency score, $S_{cls}$, for a sample is calculated using the transformed object class probability distribution $\tilde{p}_{cls}$ and the class probability distribution of the segmentation $p_{seg}$ for each pixel $i,j$ in BoxMask as follows:
\begin{multline}\label{eq:boxmask_classification}
S_{cls} = \frac{1}{2 |BM|}\sum_{i,j \in BM}KL\big(p_{seg}(i,j), \tilde{p}_{cls}\big) + \\ KL\big(\tilde{p}_{cls}, p_{seg}(i,j)\big)
\end{multline}

where $KL$ represents the Kullback-Leibler divergence between the predicted probability distributions, and the resulting value is averaged to obtain a similarity measure.

\subsection{Segmentation consistency}\label{sec:inverse_boxmask}

In \cref{sec:boxmask_localization} and \ref{sec:boxmask_classification}, we addressed inconsistencies within the boundaries of detected boxes. However, another constraint that must be met between the two tasks is that there should be no segmentation mask outside the boundaries of the detected box for a class that also belongs to the set of classes predicted in the object detection $C_{det}$.

To account for the constraint that outside of the detected boxes, there should be no segmentation mask for classes predicted in object detection, we combine all BoxMasks by taking the pixel-wise maximum and taking the inverse of the resulting binary mask to define the region falling outside of the detected objects, which we denote as $BM'$. The inconsistency score for the remaining segmented areas $S_{seg}$ is calculated using  \cref{eq:inverse_boxmask}:
\begin{equation}\label{eq:inverse_boxmask}
   S_{seg} = \frac{1}{|BM'|}\sum_{i,j \in BM'}{\mathbb{I}{(p_{seg}(i,j) \in C_{det})}}
\end{equation}

where $\mathbb{I}$ is the indicator function. For each pixel falling within the inverted BoxMask region, we penalize the class probabilities for the classes predicted by object detection. The resulting score is normalized by the number of pixels, ensuring it is scale-invariant and shares the same range as the other inconsistency scores.

\subsection{Combination of all constraints}\label{sec:combine}

The pseudo-code of combining the individual constraint scores into a single inconsistency score between the two tasks is given in \cref{alg:box_mask}. We first calculate each detected box's BoxMask as described in \cref{sec:boxmask_generation}. Then, we calculate the localization and classification consistency scores using \cref{eq:boxmask_localization} and \cref{eq:boxmask_classification}, respectively. We then add the localization and classification scores to obtain a per-box consistency score $S^{box}$. Next, as explained in \cref{sec:inverse_boxmask}, we combine all BoxMasks and traverse the inverse region to search for segmentation pixels belonging to classes from the object detection and calculate the segmentation inconsistency using \cref{eq:inverse_boxmask}. Finally, we add this score with the maximum per-box score to estimate the inconsistency between two tasks as a single selection score.

\begin{algorithm}
\caption{The pseudo-code of combining the constraints}\label{alg:box_mask}
\begin{algorithmic}[1]
\renewcommand{\algorithmicrequire}{\textbf{Input:}}
\renewcommand{\algorithmicensure}{\textbf{Output:}}
    \Require $p_{det}$, $p_{seg}$
    \Ensure score S
    \State $BM^{comb} = \{\}$
    \For{$box \in p_{det}$}
        \State Obtain $BM^{box}$ explained in \cref{sec:boxmask_generation}
        \State Compute $S_{loc}$ using \cref{eq:boxmask_localization}
        \State Compute $S_{cls}$ using \cref{eq:boxmask_classification}
        \State $S^{box} = S_{loc} + S_{cls}$
        \State $BM^{comb} = pixelwise\_max(BM^{comb}, BM^{box})$
    \EndFor
    \State $BoxMask' = inverse(BM^{comb})$
    \State Compute $S_{seg}$ using \cref{eq:inverse_boxmask}
    \State $S = S_{seg} + \underset{box \in p_{det}}{\max}(S^{box})$
\end{algorithmic}
\end{algorithm}

\section{Experiments}

\begin{figure*}[t]
  \centering
  \begin{subfigure}{0.32\linewidth}
    \includegraphics[width=0.95\linewidth]{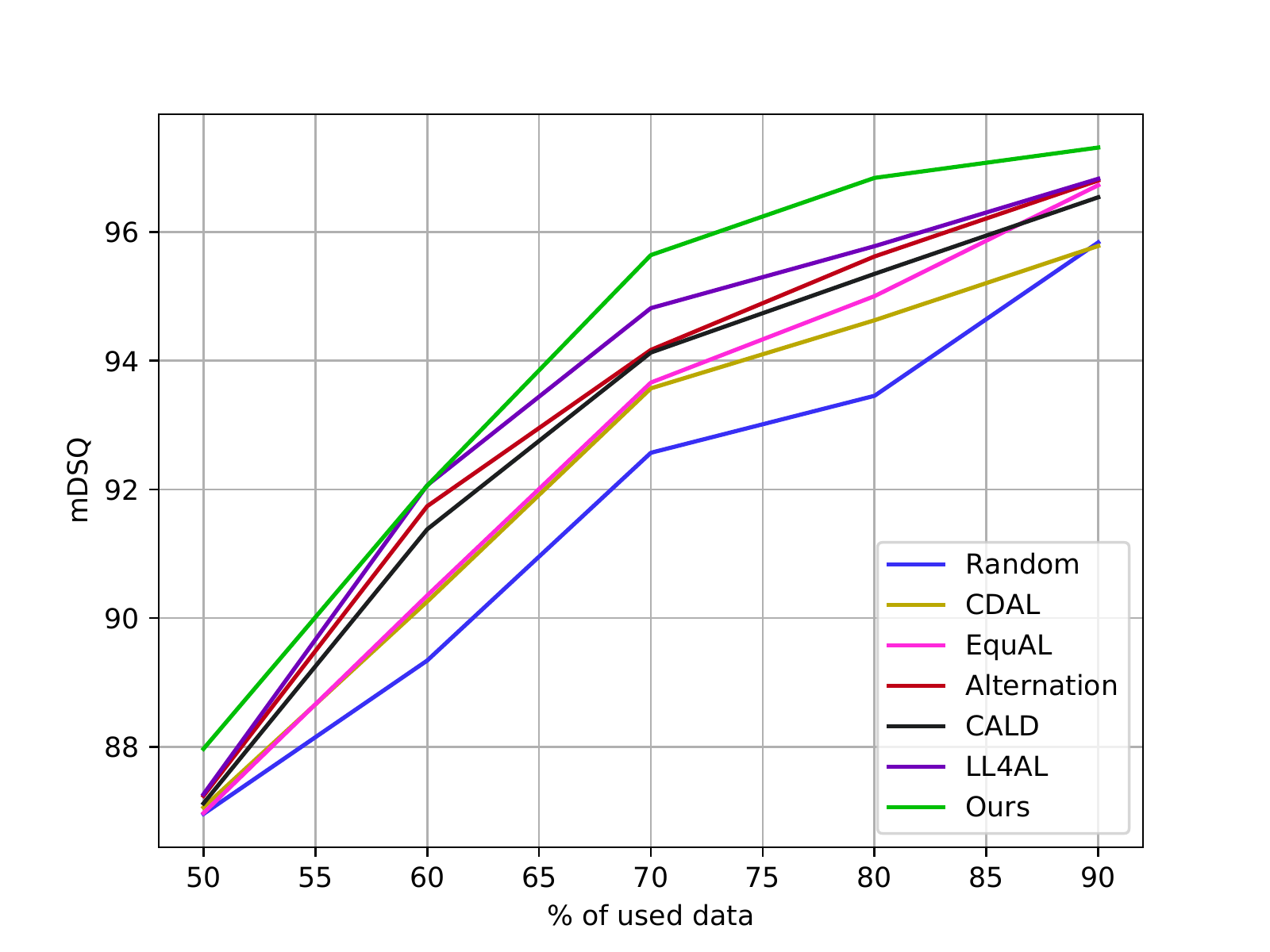}
    \caption{Multi-task performance (mDSQ)}
    \label{fig:nu_cycle-a}
  \end{subfigure}
  \hfill
  \begin{subfigure}{0.32\linewidth}
    \includegraphics[width=0.95\linewidth]{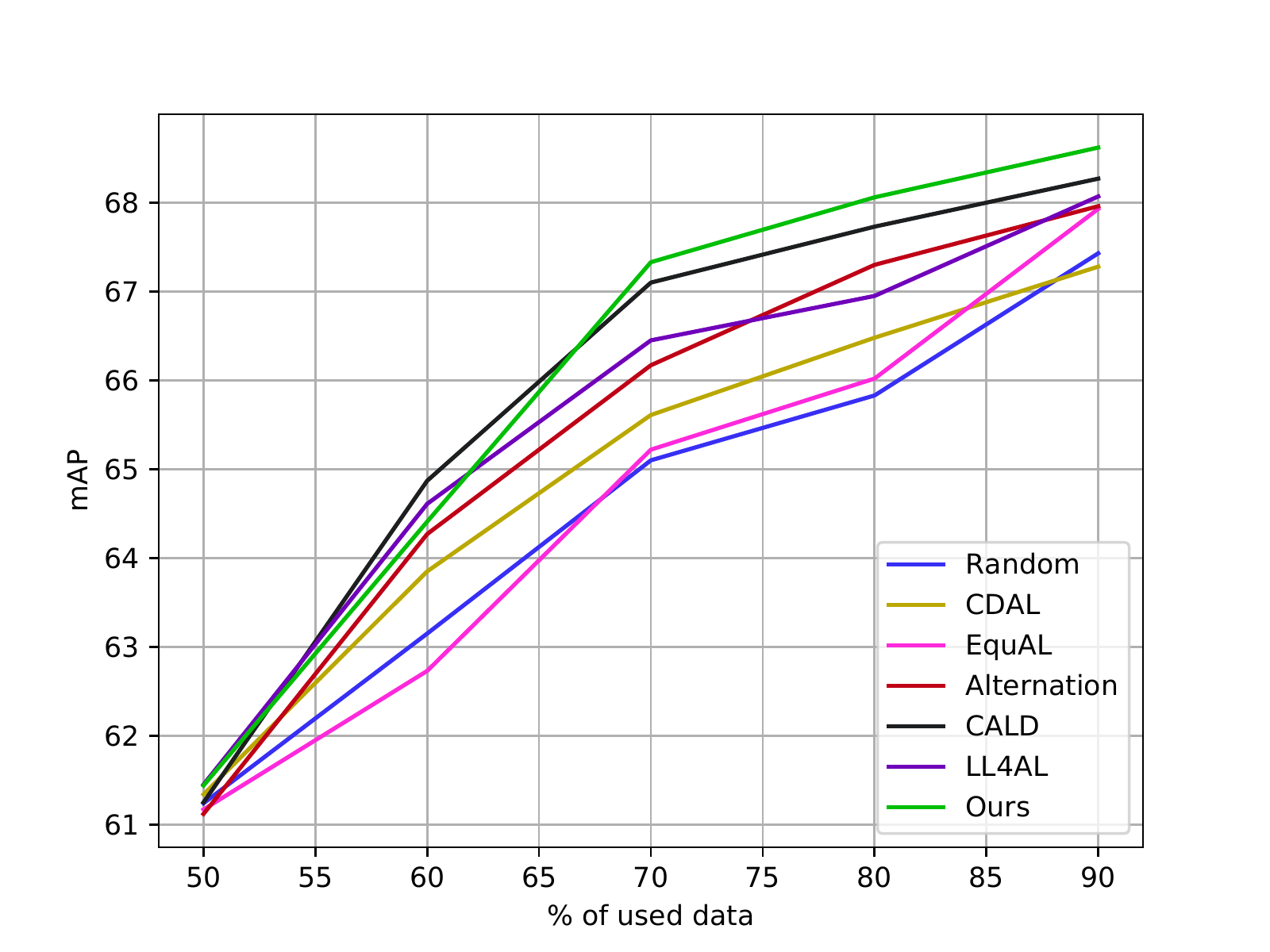}
    \caption{Detection performance (mAP)}
    \label{fig:nu_cycle-b}
  \end{subfigure}
  \hfill
  \begin{subfigure}{0.32\linewidth}
    \includegraphics[width=0.95\linewidth]{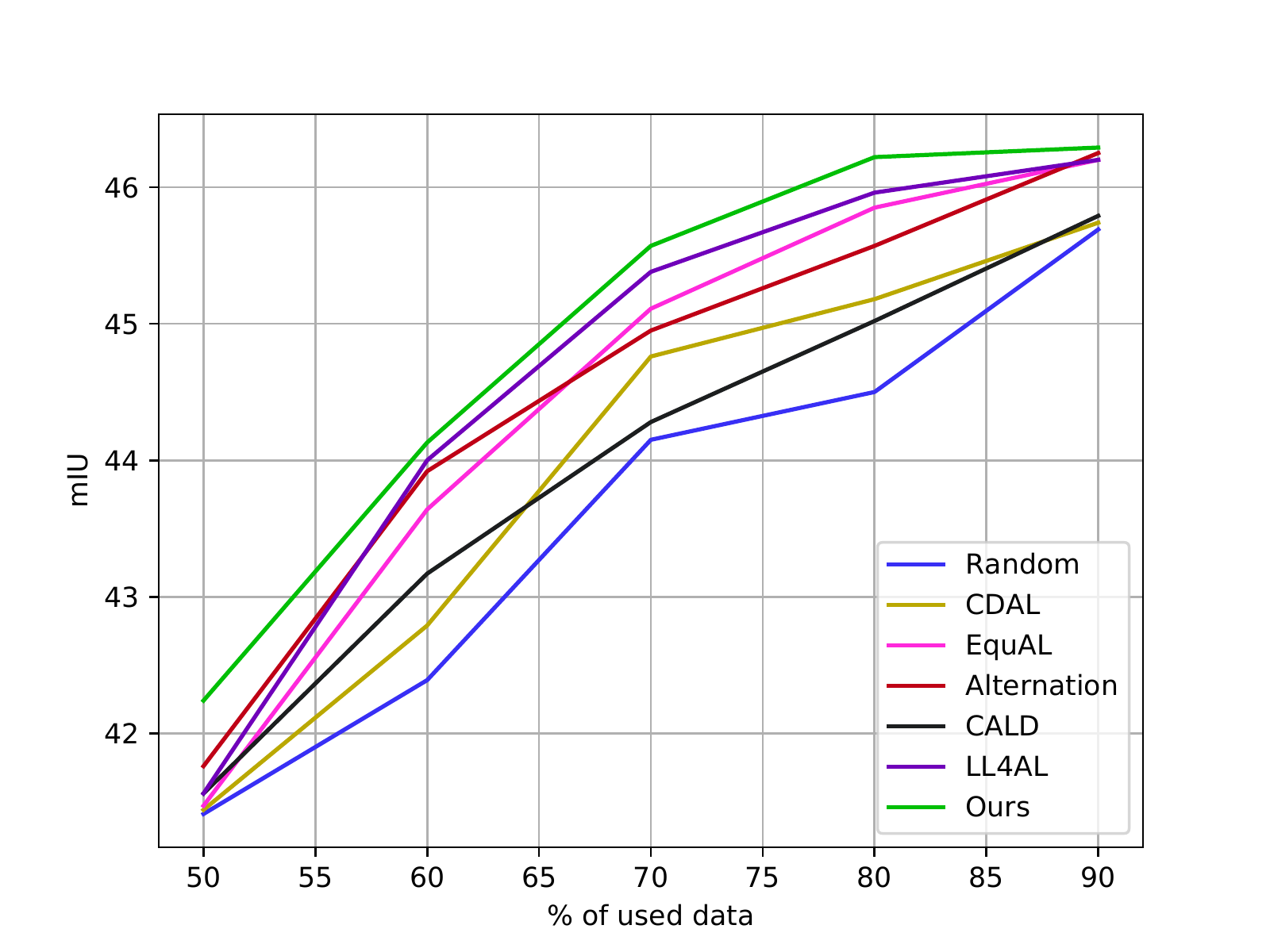}
    \caption{Segmentation performance (mIoU)}
    \label{fig:nu_cycle-c}
  \end{subfigure}
  \caption{Comparison of our proposed method with SOTA AL methods on the nuImages dataset. Lines indicate the average results over three trials. Note that all the methods start with the same network trained with 40\% of labeled samples.}
  \label{fig:nu_cycle}
\end{figure*}

\begin{figure*}[t]
  \centering
  \begin{subfigure}{0.32\linewidth}
    \includegraphics[width=0.95\linewidth]{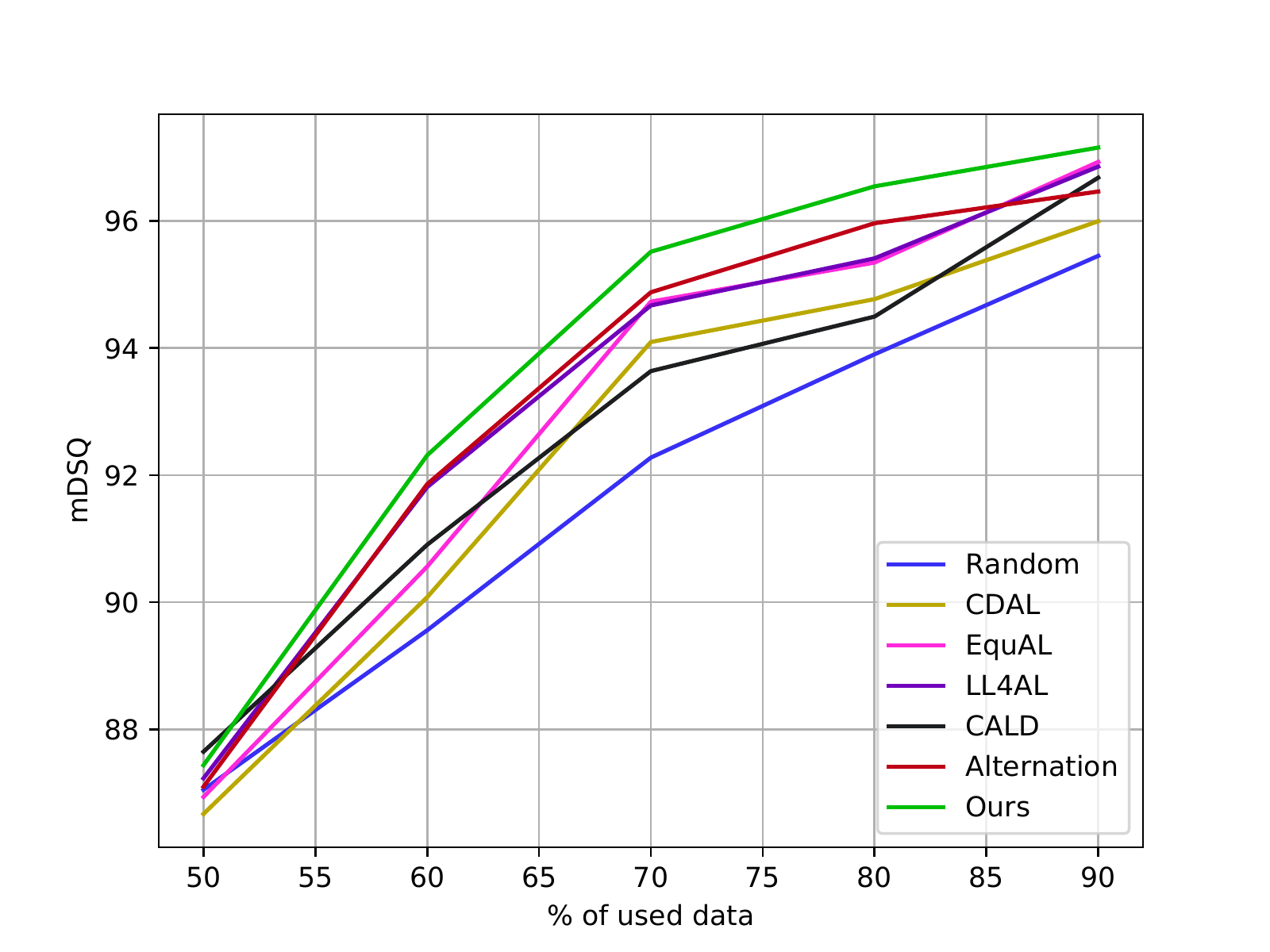}
    \caption{Multi-task performance (mDSQ)}
    \label{fig:a9_cycle-a}
  \end{subfigure}
  \hfill
  \begin{subfigure}{0.32\linewidth}
    \includegraphics[width=0.95\linewidth]{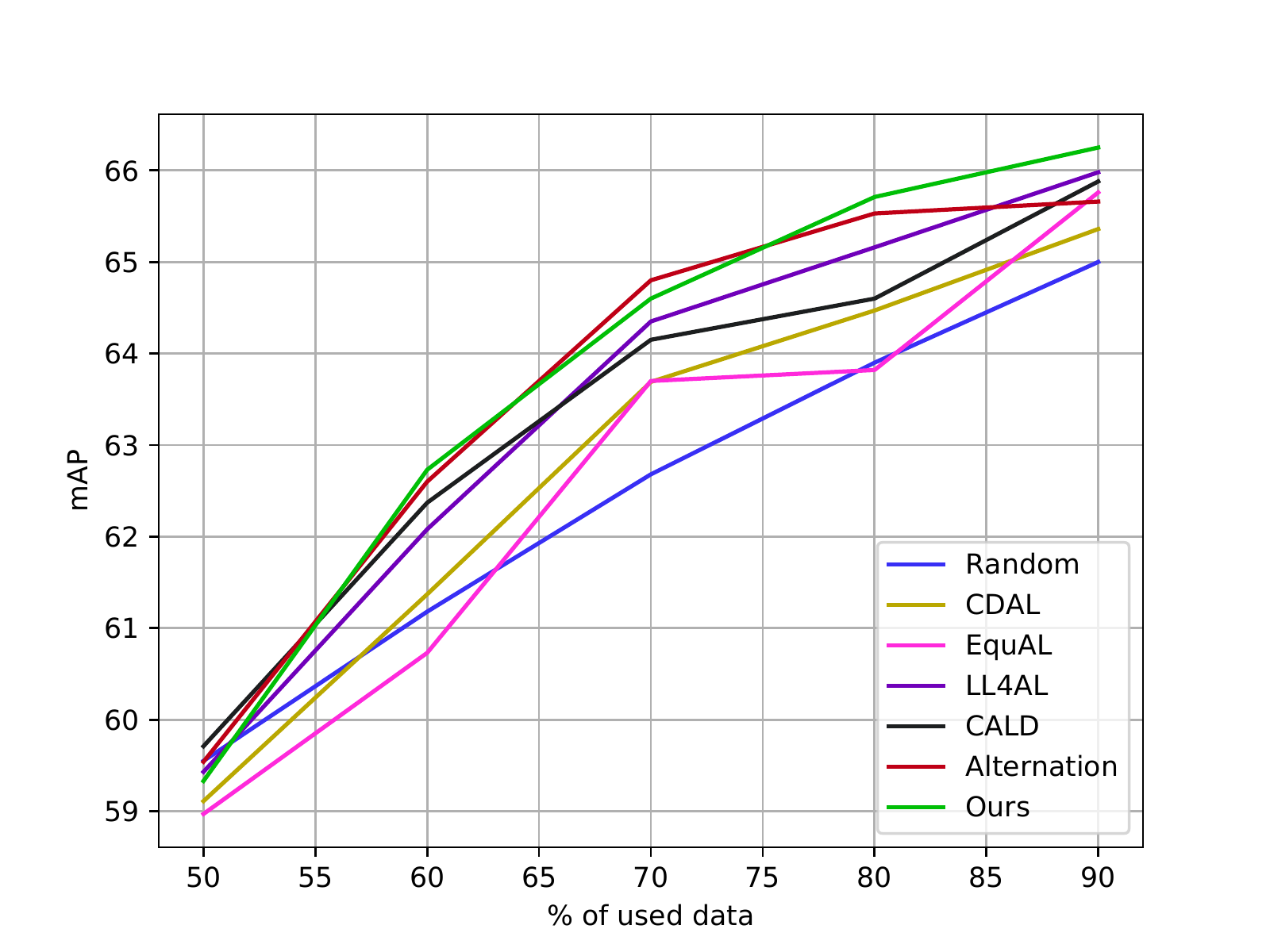}
    \caption{Detection performance (mAP)}
    \label{fig:a9_cycle-b}
  \end{subfigure}
  \hfill
  \begin{subfigure}{0.32\linewidth}
    \includegraphics[width=0.95\linewidth]{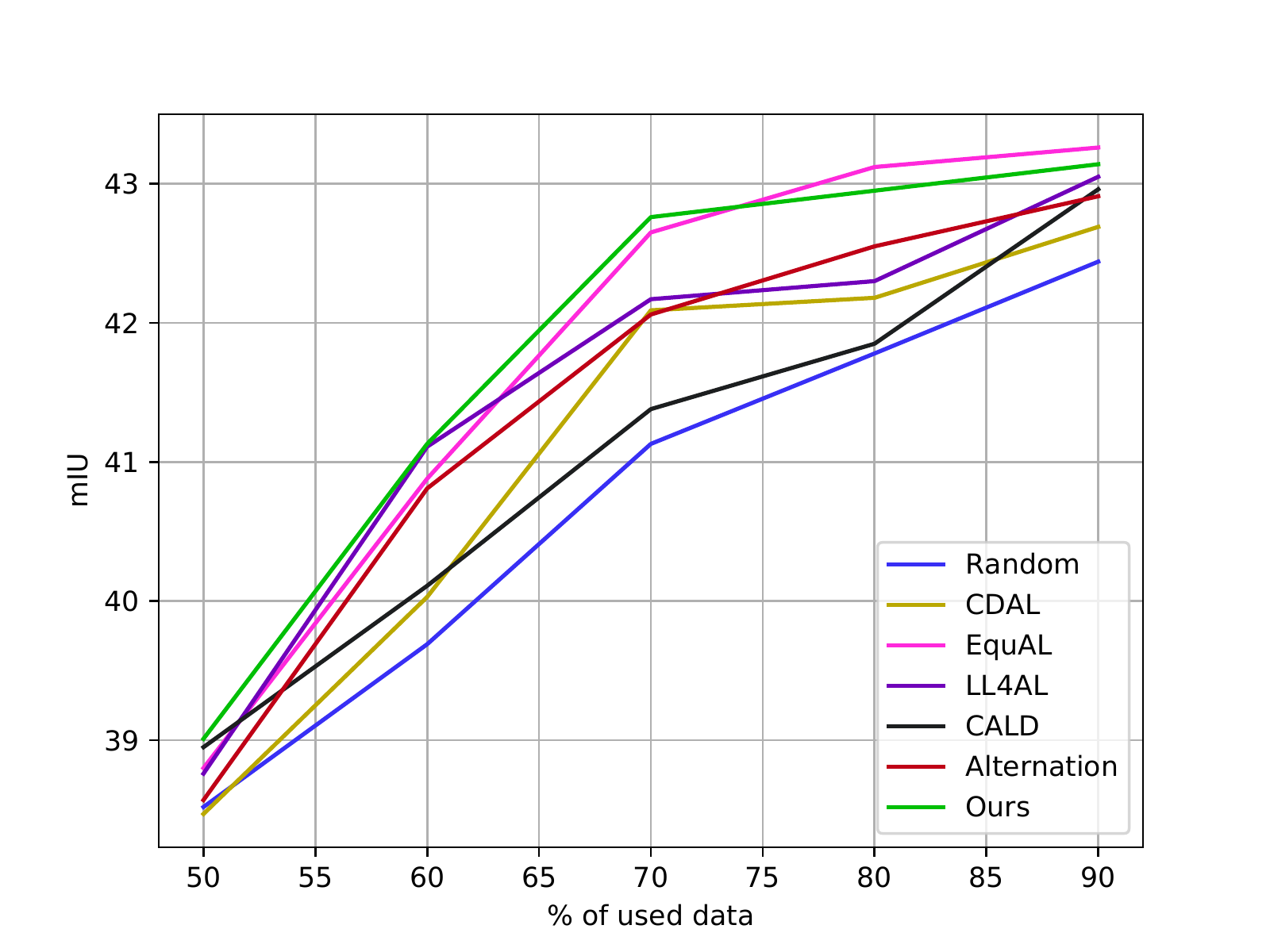}
    \caption{Segmentation performance (mIoU)}
    \label{fig:a9_cycle-c}
  \end{subfigure}
  \caption{Comparison of our proposed method with SOTA AL methods on the A9 dataset. Lines indicate the average results over three trials. Note that all the methods start with the same network trained with 40\% of labeled samples.}
  \label{fig:a9_cycle}
\end{figure*}

\subsection{Experimental setup} \label{sec:experimental_setup}

\textbf{Datasets.} We evaluate the performance of our approach on two publicly available datasets: nuImages \cite{caesar2020nuscenes} and the A9-Dataset \cite{cress2022a9}. The nuImages dataset provides 3D and 2D sensor data collected from autonomous vehicles operating in urban settings. We use images taken from the front camera, resulting in a training set size of 13,187 images and 3,249 images for the validation set with a total of 138,569 objects. The first release of the A9 dataset offers camera and LiDAR frames from two overhead gantry bridges on the A9 autobahn near Munich, Germany. It provides annotations for object detection and semantic segmentation with 33,378 labeled image frames and a total of 672,049 3D and 2D object labels. The A9 dataset was created using the \textit{proAnno} labeling toolbox which is based on \cite{zimmer20193d}.

\textbf{Implementation details.} We employ the multi-task network architecture proposed by Salscheider \etal \cite{salscheider2020simultaneous}, augmented with the loss prediction module and the active learning framework. The hyperparameters proposed in the original work are used. The experiments are performed using a batch size of 4 and a learning rate of 0.001.

To perform active learning, we randomly divide the training set into a labeled pool of 40\% and an unlabeled pool of 60\%. The initial labeled pool is used to pre-train the network, and at each iteration, the top 10\% of the samples with the highest scores are selected from the unlabeled pool to add to the labeled pool, based on the available annotations. We perform six active learning iterations of 30,000 steps per iteration for each dataset. We employ a continuous training strategy, where each active learning iteration is initialized with the best-performing checkpoint from the previous iteration. All experiments are conducted using two Tesla V100 GPUs and evaluated on the respective validation sets. 

\textbf{Evaluation metrics.} The evaluation metrics for object detection and semantic segmentation are typically measured using mean Average Precision (mAP) \cite{feng2021review} and mean Intersection-over-Union (mIoU) \cite{everingham2015pascal}, respectively. However, a new metric that can capture the performance of both tasks is necessary to evaluate the performance of multi-task active learning methods. Therefore, we propose the mean Detection Segmentation Quality (mDSQ) metric, which normalizes mAP and mIoU by the performance of the fully-trained network and combines them, as shown in \cref{eq:mDSQ}.

\begin{equation}\label{eq:mDSQ}
mDSQ = (\frac{mAP}{mAP_{fully}} + \frac{mIoU}{mIoU_{fully}})/2
\end{equation}

where $mAP_{fully}$ and $mIoU_{fully}$ represent the performance of the network trained with 100\% of data for 300,000 steps. This metric is more suitable for comparing multi-task active learning methods than the individual metrics used in each task, as it combines both metrics into a single score normalized by the fully-trained performance.

We evaluate the performance of our method using the mDSQ metric and report the mean of the metric by running three experiments with three random initial data pools. We present each experiment's numerical values and variance in the supplementary.

\subsection{Baselines}

To compare the effectiveness of our multi-task active learning method, we compare it against several baselines from the literature. We choose two inconsistency-based AL methods from the literature: we use \textit{CALD} \cite{yu2022consistency} as the SOTA method for object detection, and \textit{EquAL} \cite{golestaneh2020importance} for the semantic segmentation. We also compare against the alternating selection strategy, \textit{Alternation}, proposed by Reichart and Rappoport \cite{reichart2008multi}, and alternate between two SOTA detection and segmentation selection scores \textit{CALD} and \textit{EquAL}. Due to its task-agnostic nature, we also compare our method against the loss prediction strategy, \textit{LL4AL}, proposed by Yoo \etal \cite{yoo2019learning}. We extend the network architecture by two loss prediction modules that learn to predict the loss of each task. The loss of both tasks is then summed together to form a combined loss score. We use \textit{CDAL} \cite{agarwal2020contextual} as our diversity-based baseline, and to mimic passive learning, we use \textit{Random} selection, where each sample is assigned a score following a uniform distribution.

\subsection{Qualitative results}\label{sec:qualitative}

\begin{figure*}
    \begin{center}
    \includegraphics[width=0.9\linewidth]{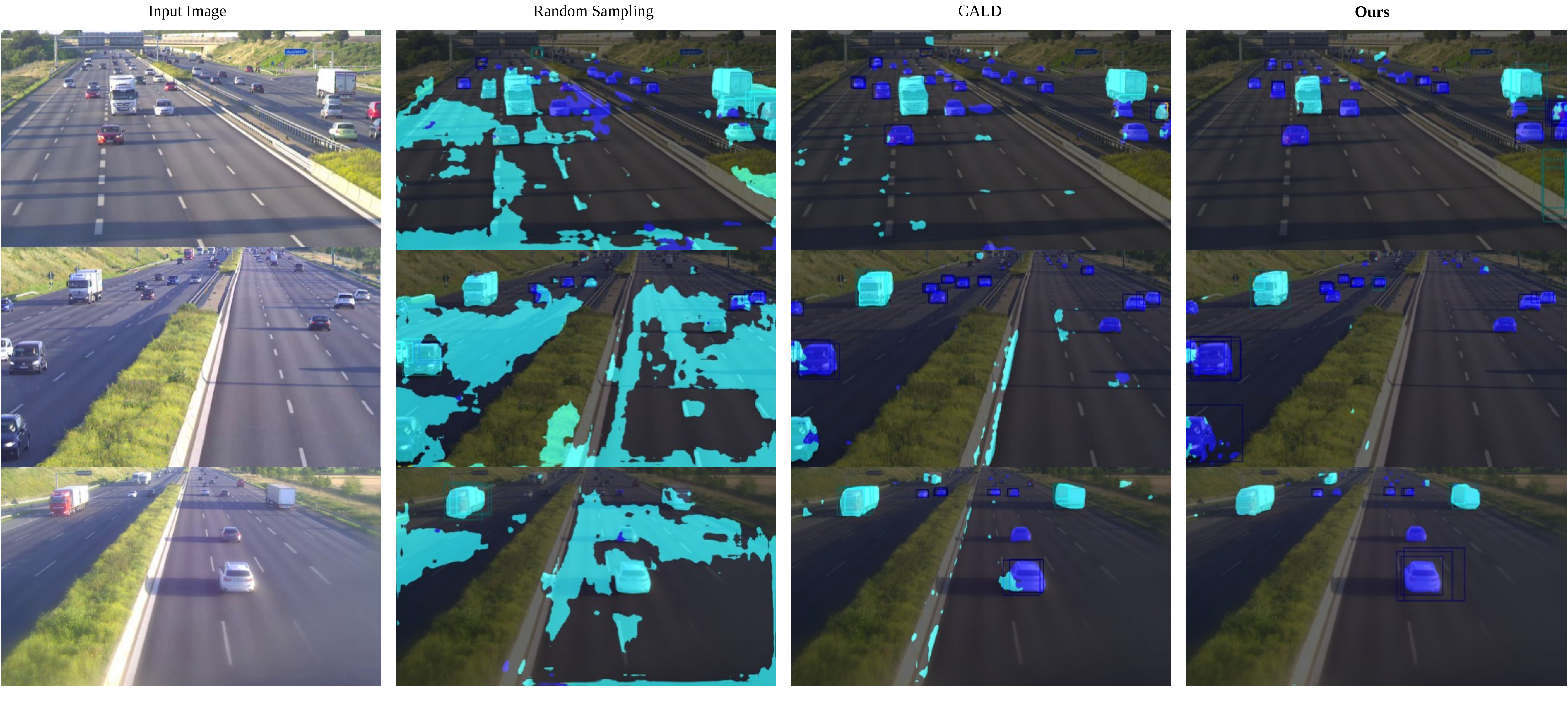}
    \end{center}
    \caption{Qualitative comparison of the Random (second), CALD (third), and Our (fourth) sampling strategies on the A9 dataset. Light blue and dark blue correspond to the \textit{Truck} and \textit{Car} classes.}
    \label{fig:qualitative}
\end{figure*}

We present qualitative results to compare the final performances obtained by following different selection strategies in Fig. \ref{fig:qualitative}. Compared to the \textit{CALD} baseline, our method provides more fitting segmentation masks within each detection. For example, for the bottom vehicle in the last row, our method correctly segments the regions as \textit{Car} instead of \textit{Truck}, demonstrating the effectiveness of using class inconsistency. Additionally, our method produces more accurate localization for the left front wheel, which is another constraint we set in our method. Finally, the pixels outside of the boxes are worse than our method, demonstrating the effectiveness of the segmentation constraint. Overall, our qualitative comparison shows that our method outperforms the baselines in producing accurate and consistent object detection and segmentation results.

\begin{table}
    \begin{center}
    \scalebox{1.0}{
    \begin{tabular}{|l|c|c|c|}
    \hline
    Score & 50\% & 70\% & 90\% \\
    \hline\hline
    $S_{loc}$ & 86.9 & 94.0 & 96.4\\
    $S_{cls}$ & 86.5 & 93.5 & 96.6\\
    $S_{seg}$ & 87.1 & 94.4 & 96.9\\
    \hline
    $S_{loc}+S_{cls}$ & 87.4 & 93.7 & 96.6\\
    $S_{cls}+S_{seg}$ & 87.0 & \underline{95.0} & 97.0\\
    $S_{loc}+S_{seg}$ & \underline{87.7} & 94.6 & \textbf{97.5}\\
    \hline
    $S_{loc}+S_{cls}+S_{seg}$ & \textbf{88.0} & \textbf{95.6} & \underline{97.3} \\
    \hline
    \end{tabular}
    }
    \end{center}
    \caption{Ablation study of the contribution of each scoring constraint for each amount of used data on nuImages.}
    \label{table:boxmask_ablation}
\end{table}

\subsection{Quantitative results} \label{sec:quantitative}

\textbf{nuImages.} Our results on the nuImages dataset are presented in \cref{fig:nu_cycle}, which shows the mDSQ, mAP, and mIoU metrics. In the initial AL cycle, our method outperforms all the baselines by at least 0.71\%. As the number of actively selected labels increases, for example, using 80\% of all available data, with 50\% actively labeled, our method outperforms Random by 3.39\% and the second-best method, \textit{LL4AL}, by 1.07\%.

Our approach reaches 95\% of the fully-trained performance using only 67\% of the data, compared to 74\% of \textit{LL4AL} and 87\% of random selection, corresponding to 20\% of more data savings. We observe that both of the multi-task selection scores (Ours, \textit{LL4AL}) outperform the single-task scores, their alternation and the diversity-based method. This demonstrates that a score that considers both tasks is more suitable for multi-task networks, compared to alternating between single-task scores as previously done.

Regarding single-task performance, as shown in \cref{fig:nu_cycle-b} and \cref{fig:nu_cycle-c}, our method is on par with the SOTA detection algorithm \textit{CALD} and even outperforms it as the number of actively selected samples increases. For semantic segmentation, our method outperforms the SOTA segmentation algorithm \textit{EquAL}. These results demonstrate that both tasks benefit from inconsistency information from the other task.

\textbf{A9.} We present the mDSQ, mAP, and mIoU metrics for the A9 dataset in \cref{fig:a9_cycle}. Our method outperforms all the baselines for all data percentages, demonstrating the effectiveness of our data selection strategy in a larger dataset. Our selection strategy has the highest performance in the first cycle, leading to the best performance throughout the remaining cycles. We achieve 95\% mDSQ using only 66\% of the data, which means a 34\% savings in labeling budget compared to full training.

For single-task detection performance (\cref{fig:a9_cycle-b}), our method is on par with the \textit{Alternation} baseline. For semantic segmentation (\cref{fig:a9_cycle-c}), our method is on par with the SOTA segmentation algorithm \textit{EquAL}. These results demonstrate that our multi-task approach can achieve comparable or better performance than state-of-the-art single-task active learning methods for both object detection and semantic segmentation.

\subsection{Ablation studies}\label{sec:ablation}

\textbf{Ablation on each component.} We perform an ablation study to evaluate the contribution of each scoring constraint function to the overall performance of our method. The results are shown in \cref{table:boxmask_ablation}. Among the single-score versions, $S_{seg}$ has the highest performance, indicating the importance of avoiding segmented pixels outside the bounding boxes. $S_{loc}$ and $S_{cls}$ have comparable results, as they are both focused on different types of inconsistency, namely classification, and localization. We observe that the two combinations lead to better performance compared to their single-score counterparts. The best performance is achieved when we combine all three scores, as shown in the last row of the table. These results demonstrate the effectiveness of each constraint and the importance of combining them to achieve optimal performance.

\begin{figure}
  \begin{center}
  \includegraphics[width=0.9\linewidth]{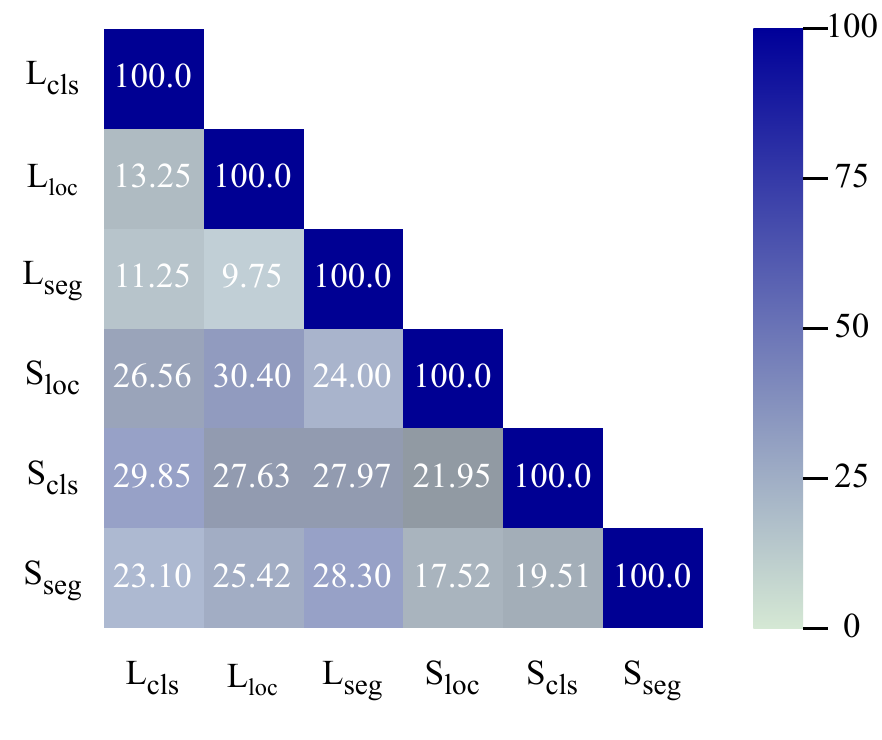}
  \end{center}
  \caption{Correlation of each constraint and losses}
  \label{fig:ablation_correlation}
\end{figure}

\textbf{Analysis of BoxMask accuracy and threshold.} The accuracy of the BoxMask is a crucial factor in obtaining accurate constraints in our method. We conduct an experiment to evaluate the accuracy of the BoxMask and compared it with using a separate network trained solely for semantic segmentation. We calculate the mIoU between the ground-truth binary segmentation label and the predict segmentation mask. We are only interested in the areas in the ground-truth segmentation that belong to the class from the detected box and only for the area bounded by the detected boxes. Based on our results, we select a threshold value $\tau$ of 0.3 for BoxMask generation. Even though our method does not have any additional parameters, it still performs comparably to using a separate network. Therefore, we use the same segmentation head from our multi-task network for generating the BoxMask predictions.

\textbf{Correlation of each consistency score with the actual error.}
We analyze the correlation of each consistency score with respect to the losses and each other, presented in \cref{fig:ablation_correlation}. We observe that $S_{seg}$ is better at measuring segmentation error, while $S_{cls}$ and $S_{loc}$ are the most effective at measuring classification and localization error in the detected boxes, respectively. Since these three losses are the main components in a joint detection and segmentation loss function, all three constraints effectively capture areas where the individual losses are high. We also observe that $S_{loc}$ and $S_{seg}$ have the lowest correlation across the selection scores, which explains the highest performance when combined in \cref{table:boxmask_ablation}.

\begin{table}
    \begin{center}
    \scalebox{1.0}{
    \begin{tabular}{|l|c|c|c|}
    \hline
    Mask Gen. & $\tau$ & Add Mem. & Accuracy \\
    \hline\hline
    \multirow{4}{*}{BoxMask} & 0.1 & - & 74.7 \\
     & 0.3 & - & \underline{78.4}\\
     & 0.5 & - & 77.9\\
     & 0.7 & - & 73.2\\
    \hline
    HR-Net & - & 6.31 GB & \textbf{80.2}\\
    \hline
    \end{tabular}
    }
    \end{center}
    \caption{Performance of BoxMask across different thresholds.}
    \label{table:boxmask_performance}
\end{table}

\begin{figure}
  \begin{center}
  \includegraphics[width=0.9\linewidth]{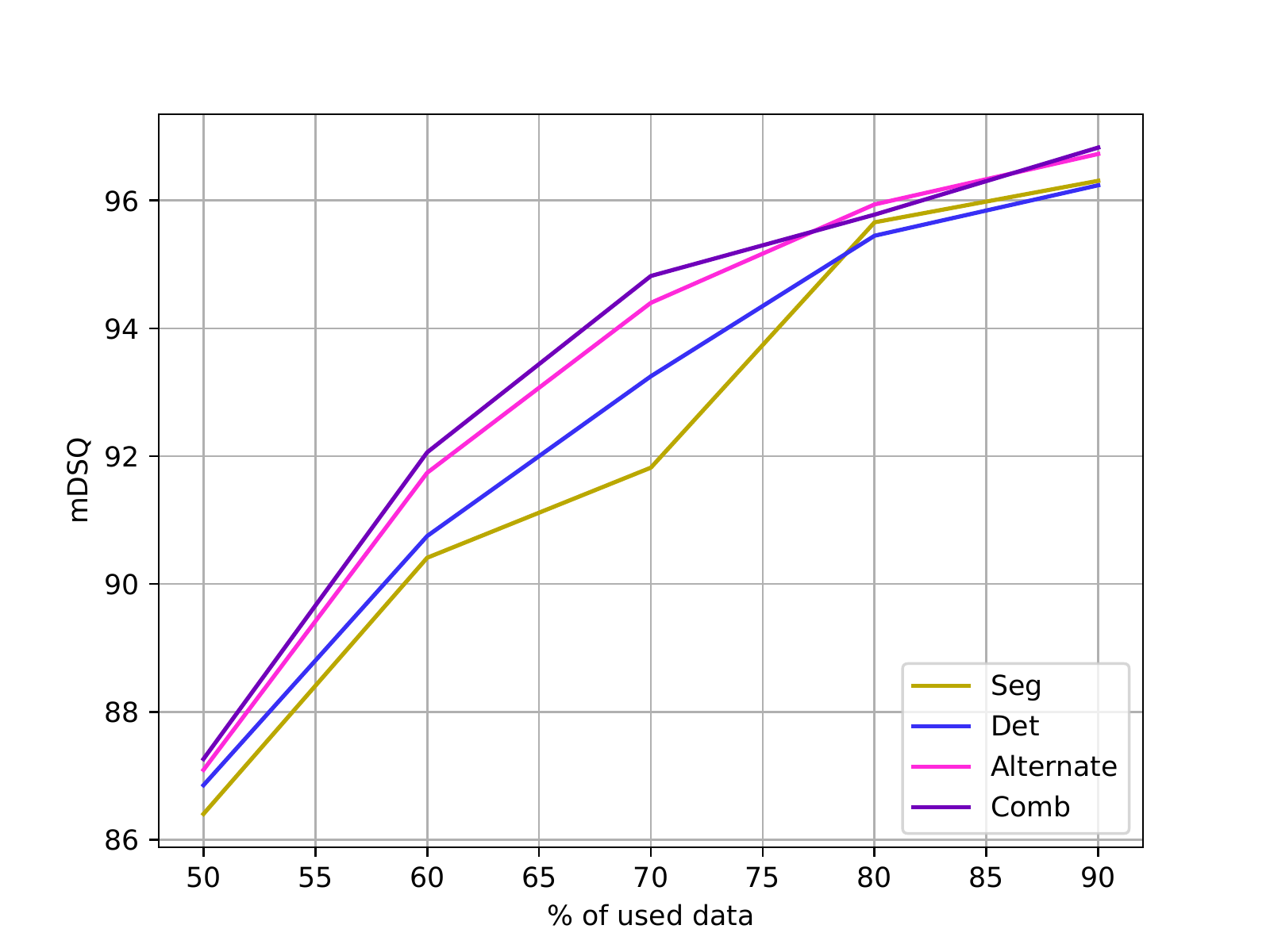}
  \end{center}
  \caption{Comparison between multi-task LL4AL to single-task losses and their alternation. \textit{Seg} and \textit{Det} refer to LL4AL using only the segmentation and detection losses, respectively. All methods start with the same network trained with 40\% of labeled data.}
  \label{fig:ablation_loss}
\end{figure}

\textbf{Single-task LL4AL and alternation.} To support our hypothesis that multi-task active learning is more effective in dealing with multiple tasks simultaneously, we compare the performance of the LL4AL strategy when using two single-task scores and one multi-task score. As shown in \cref{fig:ablation_loss}, using a single multi-task score outperforms the single-task scores and their alternation. This suggests that a score considering both tasks is more suitable for multi-task networks than alternating between single-task scores.

\section{Conclusion}

This study addressed the gap in research on active learning for multi-task networks in the vision domain. Our proposed selection strategy combines knowledge from the two task domains, object detection and semantic segmentation, into a single multi-task selection score. This strategy relies on three constraints between the two tasks and measures them by identifying the pixels belonging to a detected object through the BoxMask. Our experiments on two multi-task datasets demonstrate the effectiveness of our approach, as it outperforms all the baselines by 3.4\% and achieves 5\% fewer annotations. Future work will focus on adapting our task inconsistency-based selection strategy to other multi-task networks.









\addtolength{\textheight}{-12cm}   


\begin{thebibliography}{10}
\providecommand{\url}[1]{#1}
\csname url@samestyle\endcsname
\providecommand{\newblock}{\relax}
\providecommand{\bibinfo}[2]{#2}
\providecommand{\BIBentrySTDinterwordspacing}{\spaceskip=0pt\relax}
\providecommand{\BIBentryALTinterwordstretchfactor}{4}
\providecommand{\BIBentryALTinterwordspacing}{\spaceskip=\fontdimen2\font plus
\BIBentryALTinterwordstretchfactor\fontdimen3\font minus
  \fontdimen4\font\relax}
\providecommand{\BIBforeignlanguage}[2]{{%
\expandafter\ifx\csname l@#1\endcsname\relax
\typeout{** WARNING: IEEEtran.bst: No hyphenation pattern has been}%
\typeout{** loaded for the language `#1'. Using the pattern for}%
\typeout{** the default language instead.}%
\else
\language=\csname l@#1\endcsname
\fi
#2}}
\providecommand{\BIBdecl}{\relax}
\BIBdecl

\bibitem{liu2022swin}
Z.~Liu, H.~Hu, Y.~Lin, Z.~Yao, Z.~Xie, Y.~Wei, J.~Ning, Y.~Cao, Z.~Zhang,
  L.~Dong \emph{et~al.}, ``Swin transformer v2: Scaling up capacity and
  resolution,'' in \emph{CVPR}, 2022.

\bibitem{xu2021end}
M.~Xu, Z.~Zhang, H.~Hu, J.~Wang, L.~Wang, F.~Wei, X.~Bai, and Z.~Liu,
  ``End-to-end semi-supervised object detection with soft teacher,'' in
  \emph{ICCV}, 2021.

\bibitem{duan2019centernet}
K.~Duan, S.~Bai, L.~Xie, H.~Qi, Q.~Huang, and Q.~Tian, ``Centernet: Keypoint
  triplets for object detection,'' in \emph{ICCV}, 2019.

\bibitem{yuan2020object}
Y.~Yuan, X.~Chen, and J.~Wang, ``Object-contextual representations for semantic
  segmentation,'' in \emph{ECCV}, 2020.

\bibitem{cheng2020panoptic}
B.~Cheng, M.~D. Collins, Y.~Zhu, T.~Liu, T.~S. Huang, H.~Adam, and L.-C. Chen,
  ``Panoptic-deeplab: A simple, strong, and fast baseline for bottom-up
  panoptic segmentation,'' in \emph{CVPR}, 2020.

\bibitem{dvornik2017blitznet}
N.~Dvornik, K.~Shmelkov, J.~Mairal, and C.~Schmid, ``Blitznet: A real-time deep
  network for scene understanding,'' in \emph{ICCV}, 2017.

\bibitem{ebert2022multitask}
N.~Ebert, P.~Mangat, and O.~Wasenmuller, ``Multitask network for joint object
  detection, semantic segmentation and human pose estimation in vehicle
  occupancy monitoring,'' in \emph{IV}, 2022.

\bibitem{elezi2022not}
I.~Elezi, Z.~Yu, A.~Anandkumar, L.~Leal-Taixe, and J.~M. Alvarez, ``Not all
  labels are equal: Rationalizing the labeling costs for training object
  detection,'' in \emph{CVPR}, 2022.

\bibitem{colling2020metabox}
P.~Colling, L.~Roese-Koerner, H.~Gottschalk, and M.~Rottmann, ``Metabox+: A new
  region based active learning method for semantic segmentation using priority
  maps,'' in \emph{ICPRAM}, 2020.

\bibitem{beluch2018power}
W.~H. Beluch, T.~Genewein, A.~N{\"u}rnberger, and J.~M. K{\"o}hler, ``The power
  of ensembles for active learning in image classification,'' in \emph{CVPR},
  2018.

\bibitem{yoo2019learning}
D.~Yoo and I.~S. Kweon, ``Learning loss for active learning,'' in \emph{CVPR},
  2019.

\bibitem{roy2018deep}
S.~Roy, A.~Unmesh, and V.~P. Namboodiri, ``Deep active learning for object
  detection.'' in \emph{BMVC}, 2018.

\bibitem{desai2019adaptive}
S.~V. Desai, A.~L. Chandra, W.~Guo, S.~Ninomiya, and V.~N. Balasubramanian,
  ``An adaptive supervision framework for active learning in object
  detection,'' in \emph{BMVC}, 2019.

\bibitem{tang2021towards}
F.~Tang, D.~Wei, C.~Jiang, H.~Xu, A.~Zhang, W.~Zhang, H.~Lu, and C.~Xu,
  ``Towards dynamic and scalable active learning with neural architecture
  adaption for object detection,'' \emph{BMVC}, 2021.

\bibitem{li2021deep}
Y.~Li, B.~Fan, W.~Zhang, W.~Ding, and J.~Yin, ``Deep active learning for object
  detection,'' \emph{Information Sciences}, 2021.

\bibitem{choi2021active}
J.~Choi, I.~Elezi, H.-J. Lee, C.~Farabet, and J.~M. Alvarez, ``Active learning
  for deep object detection via probabilistic modeling,'' in \emph{ICCV}, 2021.

\bibitem{hekimoglu2022efficient}
A.~Hekimoglu, M.~Schmidt, A.~Marcos-Ramiro, and G.~Rigoll, ``Efficient active
  learning strategies for monocular 3d object detection,'' in \emph{IV}, 2022.

\bibitem{huang2021semi}
S.~Huang, T.~Wang, H.~Xiong, J.~Huan, and D.~Dou, ``Semi-supervised active
  learning with temporal output discrepancy,'' in \emph{ICCV}, 2021.

\bibitem{golestaneh2020importance}
S.~A. Golestaneh and K.~M. Kitani, ``Importance of self-consistency in active
  learning for semantic segmentation,'' in \emph{BMVC}, 2020.

\bibitem{yu2022consistency}
W.~Yu, S.~Zhu, T.~Yang, and C.~Chen, ``Consistency-based active learning for
  object detection,'' in \emph{CVPR}, 2022.

\bibitem{brust2018active}
C.-A. Brust, C.~Käding, and J.~Denzler, ``Active learning for deep object
  detection,'' in \emph{VISAPP}, 2019.

\bibitem{aghdam2019active}
H.~H. Aghdam, A.~Gonzalez-Garcia, J.~v.~d. Weijer, and A.~M. L{\'o}pez,
  ``Active learning for deep detection neural networks,'' in \emph{ICCV}, 2019.

\bibitem{yuan2021multiple}
T.~Yuan, F.~Wan, M.~Fu, J.~Liu, S.~Xu, X.~Ji, and Q.~Ye, ``Multiple instance
  active learning for object detection,'' in \emph{CVPR}, 2021.

\bibitem{kao2018localization}
C.-C. Kao, T.-Y. Lee, P.~Sen, and M.-Y. Liu, ``Localization-aware active
  learning for object detection,'' in \emph{ACCV}, 2018.

\bibitem{siddiqui2019viewal}
Y.~Siddiqui, J.~Valentin, and M.~Nie{\ss}ner, ``Viewal: Active learning with
  viewpoint entropy for semantic segmentation,'' in \emph{CVPR}, 2020.

\bibitem{li2020uncertainty}
B.~Li and T.~Alstr{\o}m, ``On uncertainty estimation in active learning for
  image segmentation,'' in \emph{ICMLW}, 2020.

\bibitem{xie2020deal}
S.~Xie, Z.~Feng, Y.~Chen, S.~Sun, C.~Ma, and M.~Song, ``Deal: Difficulty-aware
  active learning for semantic segmentation,'' in \emph{ACCV}, 2020.

\bibitem{casanova2020reinforced}
A.~Casanova, P.~O. Pinheiro, N.~Rostamzadeh, and C.~J. Pal, ``Reinforced active
  learning for image segmentation,'' in \emph{ICLR}, 2020.

\bibitem{kasarla2019region}
T.~Kasarla, G.~Nagendar, G.~M. Hegde, V.~Balasubramanian, and C.~Jawahar,
  ``Region-based active learning for efficient labeling in semantic
  segmentation,'' in \emph{WACV}, 2019.

\bibitem{xie2022towards}
B.~Xie, L.~Yuan, S.~Li, C.~H. Liu, and X.~Cheng, ``Towards fewer annotations:
  Active learning via region impurity and prediction uncertainty for domain
  adaptive semantic segmentation,'' in \emph{CVPR}, 2022.

\bibitem{sener2018active}
O.~Sener and S.~Savarese, ``Active learning for convolutional neural networks:
  A core-set approach,'' in \emph{ICLR}, 2018.

\bibitem{agarwal2020contextual}
S.~Agarwal, H.~Arora, S.~Anand, and C.~Arora, ``Contextual diversity for active
  learning,'' in \emph{ECCV}, 2020.

\bibitem{crawshaw2020multitask}
M.~Crawshaw, ``Multi-task learning with deep neural networks: A survey,''
  \emph{arXiv preprint arXiv:2009.09796}, 2020.

\bibitem{zhao2018modulation}
X.~Zhao, H.~Li, X.~Shen, X.~Liang, and Y.~Wu, ``A modulation module for
  multi-task learning with applications in image retrieval,'' in \emph{ECCV},
  2018.

\bibitem{liu2019end}
S.~Liu, E.~Johns, and A.~J. Davison, ``End-to-end multi-task learning with
  attention,'' in \emph{CVPR}, 2019.

\bibitem{ghiasi2021multitask}
G.~Ghiasi, B.~Zoph, E.~D. Cubuk, Q.~V. Le, and T.-Y. Lin, ``Multi-task
  self-training for learning general representations,'' in \emph{ICCV}, 2021.

\bibitem{gong2019comparison}
T.~Gong, T.~Lee, C.~Stephenson, V.~Renduchintala, S.~Padhy, A.~Ndirango,
  G.~Keskin, and O.~H. Elibol, ``A comparison of loss weighting strategies for
  multi task learning in deep neural networks,'' \emph{IEEE Access}, 2019.

\bibitem{kendall2017multitask}
A.~Kendall, Y.~Gal, and R.~Cipolla, ``Multi-task learning using uncertainty to
  weigh losses for scene geometry and semantics,'' in \emph{CVPR}, 2018.

\bibitem{liu2019multi}
X.~Liu, P.~He, W.~Chen, and J.~Gao, ``Multi-task deep neural networks for
  natural language understanding,'' in \emph{ACL}, 2019.

\bibitem{hariharan2014simultaneous}
B.~Hariharan, P.~Arbel{\'a}ez, R.~Girshick, and J.~Malik, ``Simultaneous
  detection and segmentation,'' in \emph{ECCV}, 2014.

\bibitem{papadopoulos2017extreme}
D.~P. Papadopoulos, J.~R. Uijlings, F.~Keller, and V.~Ferrari, ``Extreme
  clicking for efficient object annotation,'' in \emph{ICCV}, 2017.

\bibitem{ren2015faster}
S.~Ren, K.~He, R.~Girshick, and J.~Sun, ``Faster r-cnn: Towards real-time
  object detection with region proposal networks,'' in \emph{NeurIPS}, 2015.

\bibitem{salscheider2020simultaneous}
N.~O. Salscheider, ``Simultaneous object detection and semantic segmentation,''
  in \emph{ICPRAM}, 2020.

\bibitem{reichart2008multi}
R.~Reichart, K.~Tomanek, U.~Hahn, and A.~Rappoport, ``Multi-task active
  learning for linguistic annotations,'' in \emph{ACL}, 2008.

\bibitem{ikhwantri2018multi}
F.~Ikhwantri, S.~Louvan, K.~Kurniawan, B.~Abisena, V.~Rachman, A.~F. Wicaksono,
  and R.~Mahendra, ``Multi-task active learning for neural semantic role
  labeling on low resource conversational corpus,'' in \emph{ACLW}, 2018.

\bibitem{caesar2020nuscenes}
H.~Caesar, V.~Bankiti, A.~H. Lang, S.~Vora, V.~E. Liong, Q.~Xu, A.~Krishnan,
  Y.~Pan, G.~Baldan, and O.~Beijbom, ``nuscenes: A multimodal dataset for
  autonomous driving,'' in \emph{CVPR}, 2020.

\bibitem{cress2022a9}
C.~Cre{\ss}, W.~Zimmer, L.~Strand, M.~Fortkord, S.~Dai, V.~Lakshminarasimhan,
  and A.~Knoll, ``A9-dataset: Multi-sensor infrastructure-based dataset for
  mobility research,'' in \emph{IV}, 2022.

\bibitem{zimmer20193d}
W.~Zimmer, A.~Rangesh, and M.~Trivedi, ``3d bat: A semi-automatic, web-based 3d
  annotation toolbox for full-surround, multi-modal data streams,'' in
  \emph{IV}, 2019.

\bibitem{feng2021review}
D.~Feng, A.~Harakeh, S.~L. Waslander, and K.~Dietmayer, ``A review and
  comparative study on probabilistic object detection in autonomous driving,''
  in \emph{ITS}, 2021.

\bibitem{everingham2015pascal}
M.~Everingham, S.~Eslami, L.~Van~Gool, C.~K. Williams, J.~Winn, and
  A.~Zisserman, ``The pascal visual object classes challenge: A
  retrospective,'' \emph{IJCV}, 2015.

\end{thebibliography}
\end{document}